# A Mobile App for Wound Localization using Deep Learning

D M Anisuzzaman[1], Yash Patel[1], Jeffrey Niezgoda[2], Sandeep Gopalakrishnan[3*] and Zeyun Yu[1, 4*]

*Abstract*—We present an automated wound localizer from 2D wound and ulcer images by using deep neural network, as the first step towards building an automated and complete wound diagnostic system. The wound localizer has been developed by using YOLOv3 model, which is then turned into an iOS mobile application. The developed localizer can detect the wound and its surrounding tissues and isolate the localized wounded region from images, which would be very helpful for future processing such as wound segmentation and classification due to the removal of unnecessary regions from wound images. For Mobile App development with video processing, a lighter version of YOLOv3 named tiny-YOLOv3 has been used. The model is trained and tested on our own image dataset in collaboration with AZH Wound and Vascular Center, Milwaukee, Wisconsin. The YOLOv3 model is compared with SSD model, showing that YOLOv3 gives a mAP value of 93.9%, which is much better than the SSD model (86.4%). The robustness and reliability of these models are also tested on a publicly available dataset named Medetec and shows a very good performance as well.

## I. INTRODUCTION

CURRENTLY there are approximately 451 million people affected by diabetes, and this number is expected to increase to 693 million by the year of 2045 [1]. It is reported that 15% of diabetic patients are likely to develop a diabetic foot ulceration (DFU) during their lifetime [2]. Venous leg ulcer (VLU) is another major wound type. Approximately 5-8% of the world population suffers from venous disease, and 1% are likely to develop a VLU [3][4]. Pressure ulcer (PU) is another major class of wounds, responsible for a high mortality rate (29%) especially for elderly people [5]. Due to unbalanced global economic development, a large portion of wound patients in developing countries or rural regions do not have access to proper diagnostic, evidenced based treatment guidelines, appropriate technologies or clinical expertise required for optimal healing outcomes. The creation of an intelligent wound system can be beneficial in many ways including reducing clinical workloads, cost efficacy, standardizing treatment, and improving patient care. With the development of remote telemedicine systems equipped with intelligent wound analysis, patients in distant locations can have access to improved diagnostic and management strategies. A critical first step in the development of this intelligent system is wound localization. This is required to detect regions of interest (i.e., wound regions) from 2D wound images so that the subsequent processing algorithms, such as wound segmentation, classification, measurement, tissue composition analysis, can be accomplished and then integrated into the intelligent wound analysis system. To this end, we have developed a wound localization algorithm by using deep learning model and integrated it into a smartphone platform, which can remotely capture and localize a wound from 2D wound images.

Specifically, wound localization begins with placing a bounding box around the wound or ulcer in a wound image and then cropping the bounded box for further processing. Fig. 1. provides a brief overview of an automated wound analysis system. The extracted (cropped) region of a wound image will be passed as input to the segmentation and classification modules (named wound segmenter and wound classifier respectively). The wound segmenter will segment a wound image for feature quantification (area, perimeter, width, height etc.) and the wound classifier will classify the image into different types of wounds (DFU, PU, VLU etc.). The classifier also classifies wound images into different tissue composition types (such as necrotic, slough, granulation, epithelium, and healed dermis) based on the pixel colors. Wound localization can significantly simplify these subsequent wound analysis steps due to the removal of unnecessary areas of wound images. Additionally, by limiting data capture to only the wounded tissue all distinguishing features (face, tattoo, birthmark etc.) are removed thereby enhancing patient privacy via wound localization.

Previous research on wound system development identified some controlled environment (devices) used to acquire a wound image including photographic foot imaging device (PFID) [6], wound measurement device (WMD) [7], wound image capture box (WICB) [8]. Though these devices can

[1]Department of Computer Science, University of Wisconsin-Milwaukee, Milwaukee, WI, USA;

[2]Advancing the Zenith of Healthcare (AZH) Wound and Vascular Center, Milwaukee, WI, USA;

[3]College of Nursing, University of Wisconsin Milwaukee, Milwaukee, WI, USA;

4Department of Biomedical Engineering, University of Wisconsin-Milwaukee, Milwaukee, WI, USA.

*Corresponding authors:

Zeyun Yu, Associate Professor and Director of Big Data Analytics and Visualization Laboratory, Department of Computer Science, University of Wisconsin-Milwaukee, Milwaukee, WI, USA. Email: yuz@uwm.edu

Sandeep Gopalakrishnan, Assistant Professor, College of Nursing, University of Wisconsin Milwaukee, Milwaukee, WI, 53211, USA. Email: sandeep@uwm.edu

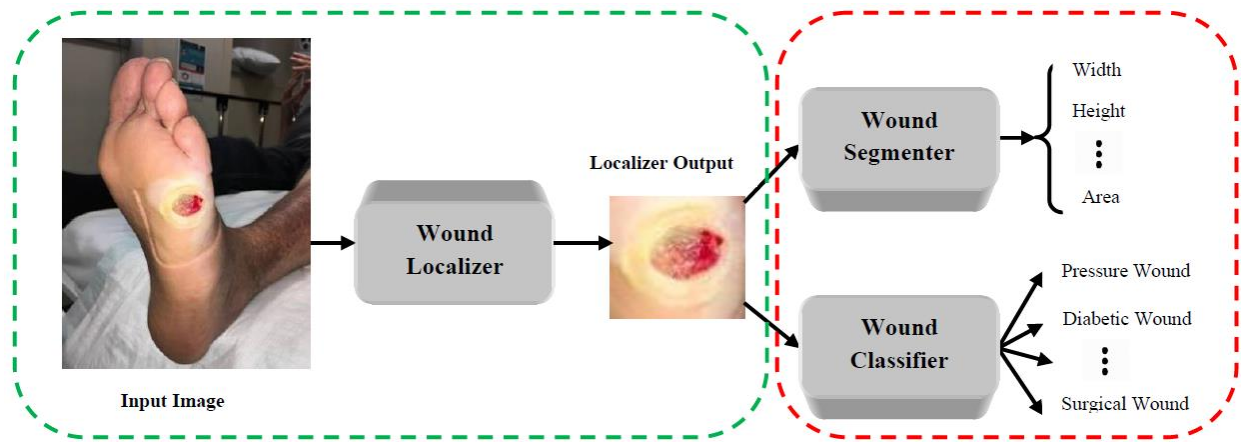

Fig. 1. Automated Wound System Overview

overcome the problem of illumination, they are expensive and often require professional operation, thereby limiting home use and preventing easy patient access and adoption. Using smartphones patients can easily capture wound images anywhere and anytime. Oftentimes smartphone imagery will capture a lot of unnecessary data. An example of extraneous data collection is demonstrated in Fig. 1., in which the wound image also captures the bed sheet, pillow, calendar, fingers, etc. Non-wound image data can confuse the wound segmentation and classification algorithms resulting in reduced performance and weakened overall performance of the wound analysis intelligent system.

One of most popular deep neural networks for image processing is convolutional neural network (CNN). In CNN, the input is presented as a tensor as a shape of "number of images × width × height × depth". For a single image with 1000 pixels in both height and width, and 3 channels (RGB) in depth, the input size to the CNN is 3 million. If the first layer of the CNN has 1000 neurons, the requirement would be to train 1 billion perimeters for this single layer CNN. Thus, it is prohibitively expensive to train these networks given the limited hardware capacity. For this reason, we have to down-sample the input, which leads to loss of information and ultimately poor performance of the network. Using our wound localizer, we provide the segmenter and classifier much smaller images which results in improved performance at significantly less cost.

Though some researchers used localized images (cropped images or patches) to train their classifier or segmenter, the wound regions are selected manually in most of the cases. To calculate the surface area of chronic wound, Papazoglou et al. [9] developed an algorithm based on the color difference of wound areas and non-wound areas. In the first step of their algorithm, they have manually selected the region of interest (ROI). The ROI was selected so that the wound is centered and constitutes most of the cropped image. Hettiarachchi et al. [14] developed an android mobile application for wound segmentation, where they applied cropping on the original image for centering the wound. Image cropping was done manually by setting up the wound boundary through selecting the diagonal points of a rectangle (bounding box). This process is claimed to be able to remove unnecessary information like clothing, limb borders and backgrounds. Chang et al. [11] developed a multimodal sensor system for wound segmentation, tissue classification, 3D wound size measurement (length, width, depth, surface, volume), thermal profiling (blood-flow and metabolic activities), multi-spectral analysis (oxygen saturation), and chemical sensing (skin odor measurement). They passed a manually selected region of interest (containing only the wound and its surrounding part) from the original image to their segmenter, classifier and other modalities [11].

Wantanajittikul et al. [10] detected the degree of burn from five burn images collected from Department of Medical Services, Ministry of Public Health, Thailand; by using SVM, K-mean and Bayes classifier. For the burn degree classification, instead of using the whole image, they fed their network with 34 sub-images of 40 × 40 pixels. These sub-images were cropped manually from the original images and two experts labeled each sub-image to its degree of burn respectively. Goyal et al. [12] developed a novel CNN architecture named DFUNet, for binary classification of healthy skin versus diabetic foot ulcer from RGB color images. They used two types of patches (healthy and ulcer), manually labeled by medical experts, as the input of their convolutional layer. An open source annotator named manual whisker annotator (MWA) [13] was employed to outline these patches from an original image. Shenoy et al. [15] proposed a CNN-based method for binary classification (positive and negative) of nine different types of wound images. They used a modified version of VGG16 network, named WoundNet, as the classifier. Their training dataset contains 1,335 wound images, where they anonymized all the images and cropped them into squares of same size. Alzubaidi et al. [16] proposed a novel deep convolutional neural network, named DFU_QUTNet, for binary patch classification of normal skin versus abnormal skin (diabetic ulcer). They cropped a significant region around the ulcer which includes important tissues of both classes. A medical specialist labeled the cropped patches into normal and abnormal classes including 542 normal and 1067 abnormal (DFU) patches. The selection and cropping of patches from original image were done manually. Pinero et al. [17] classified burn depths into five classes, superficial dermal (blisters), superficial dermal (red), deep dermal, full-thickness (beige), and full-thickness (brown), based on wound image color and texture features. They selected six features (lightness, hue, SD of hue, u* chrominance, SD of v*, and skewness of lightness) by using the sequential backward selection (SBS) method and fed them to the Fuzzy-ARTMAP neural network for the five-class classification. Instead of using the original image, they used $49 \times 49$ burn image patches to train their classifier. They have a total of 250 patches, with each class containing 50 burn image patches. These $49 \times 49$ burn patches were extracted manually, and they only contain the burned skins and exclude the healthy skins and background.

Goyal et al. [25] proposed methods for Diabetic Foot Ulcer (DFU) detection and localization on mobile devices. They introduced a dataset including 1,775 DFU images and used SSD-MobileNet, SSD-InceptionV2, Faster R-CNN with InceptionV2, and R-FCN with Resnet 101 models for wound localization. For evaluating localization performance, they used mean average precision (mAP) and overlap percentage metrices. From mAP point of view the best results were generated by Faster R-CNN with InceptionV2 model, and from overlap percentage point of view, the R-FCN with ResNet101 generates the best results. For smartphone application they used the Faster R-CNN with InceptionV2 model on an Android phone. In another work, Goyal et al. [24] proposed a new dataset of DFUs as well as a classification method that predicts the presence of infection or ischemia in the DFU. For these experiments they introduced a dataset including 1,459 DFU images. In their data augmentation step, they used Faster-RCNN and InceptionResNetV2 architectures for ROI detection. No evaluation metric was presented for wound ROI detection on their dataset.

Keeping very small amount of work on automated wound localization and its great benefits in mind, we have developed our wound localizer by using deep neural networks and further compacted our wound localizer for mobile platform. The rest of the paper is structured in the following way: Section II focuses on materials (dataset, equipment etc.) and methods; Section III briefly plots the mobile application platform; Section IV discusses the results and findings; and finally, we conclude the paper with future work in Section V.

## II. WOUND FEATURES

### A. Data Collection

The wound dataset has been collected from AZH Wound and Vascular Center, Milwaukee, WI, USA. This dataset (AZH Wound Database) contains a total of 1,010 wound images. Three types of ulcers have been included in the dataset: Diabetic foot ulcer (DFU), Pressure Ulcer (PU), and Venous Ulcer (VU). All the images are captured with iPad and DSLR cameras. No specific environmental or illumination condition has been applied during image capturing. These images are further processed and used as training and test data.

Additionally, for testing the robustness and reliability of our models, 52 images have been downloaded from Medetec Wound Database [18]. Though this database contains all types of open wound images such as abdominal wounds, burn and scalds, diabetic foot ulcers, haemangiomas, venous ulcers and arterial ulcers, malignant wounds etc., we have only collected diabetic foot ulcer, venous ulcer and pressure ulcer images due to the types of training images.

### B. Data Preparation

As our models can take different width-height ratios of images in both training and test datasets; we do not make the image size uniform. To increase the number of images for our dataset, we have applied some augmentations, on the AZH Wound Database with rotation, flipping (up and right), and blurring augmentations, which results in a total of 4,050 image data. We have used 3,645 images as training dataset and 405 images as test dataset. All the collected images have been labeled manually for training and for evaluation of our models. We have used an MIT licensed free graphical image annotation tool, named labelImg [19] for data labeling. Annotations are saved as YOLO format as a text file for each image; containing the class number, center coordinates of bounding box(s), and height and width of bounding box(s). Annotations are further changed to Pascal VOC format, and together with images, passed as the inputs to SSD model.

### C. Models

We used YOLOv3 and SSD as our wound localization models. These models are chosen for their popularity, reliability and time management for object detection. A comparison of these two models for our wound detection task has been presented in the result and discussion section. A brief description of these models is given below.

#### 1) You Only Look Once (YOLOv3)

YOLOv3 is the third generation of the YOLO family, which can predict both bounding boxes and classify the object within the bounding box in one pass. YOLOv3 does prediction on the per-frame basis and no temporal information is employed. This architecture consists of three different types of networks: Darknet-53, upsampling, and YOLO layers or detection layers. The darknet-53 network is used to extract features from the input image, consisting of residual blocks as the basic component. Each residual block consists of a pair of $3 \times 3$ and $1 \times 1$ convolutional layer together with shortcut connections. As the name suggests, there are 53 convolutional layers in Darknet-53. In the upsampling layers, YOLOv3 have a total 106 fully convolutional layers. The YOLO layers are responsible for detecting objects at different scales using features extracted by Darknet-53 layers. At the initial YOLO layer, the grid size is 1/32 of the input image size and at the final YOLO layer the grid size is 1/8 of the input image size. With three YOLO layers smaller objects can also be detected. Each YOLO layer consists of a few convolution layers with batch normalization and leaky ReLU activation. There are shortcut connections that connect darknet-53 intermediate layers to the layer after upsampling layer [20]. The model architecture of YOLOv3 is shown in Fig. 2.

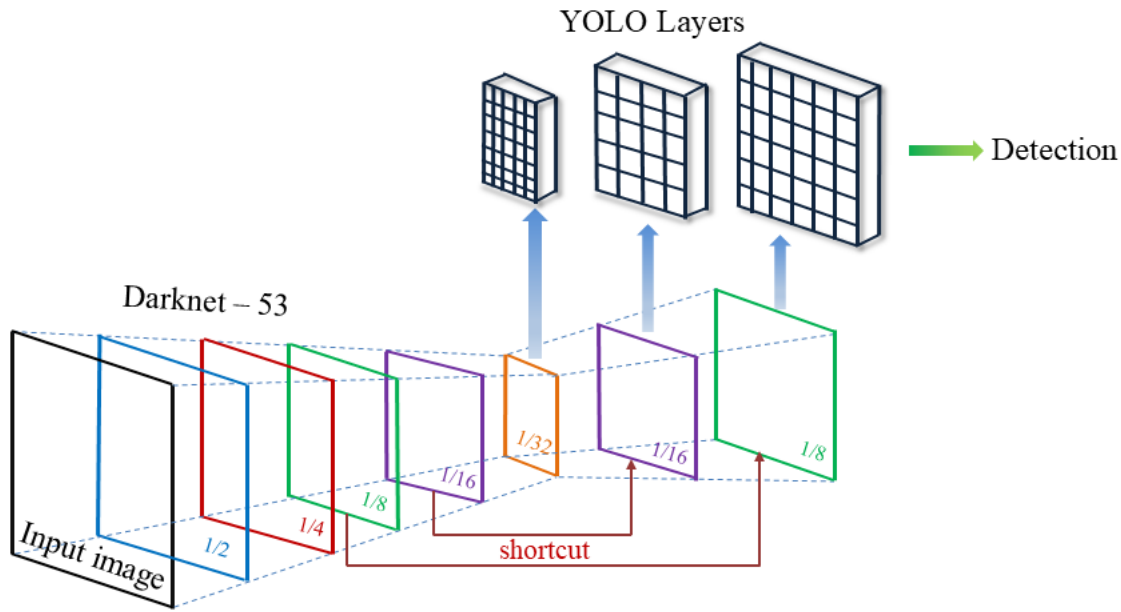

Fig. 2. YOLOv3 Network Architecture.

*2) Single Shot MultiBox Detector (SSD)*

SSD [21] stands for Single Shot MultiBox Detector, composed of mainly two object detection parts: feature maps extraction and object detection by applying convolution filters. VGG16 and Conv4_3 layer have been used for feature maps extraction and object detection respectively. For each location, it makes 4 object predictions, where each prediction consists of a boundary box and scores for each class (including the class for no object), and the highest score is selected as the class for the bounded object. After extracting the feature maps, $3 \times 3$ convolution filters are applied for each cell to make predictions. Six more auxiliary convolution layers are added after the VGG16, five of which are used for the object detection. From six layers SSD makes a total of 8,732 predictions per class, followed by a non-maximum suppression step to produce the final detections.

D. *Model Training*

We used a single class, named "wound" for our model training. For YOLOv3 model, we used the YOLO annotations. This model is trained for 273 epochs with a batch size of 8. YOLOv3 model was trained with a learning rate of 0.001, and stochastic gradient descent (SGD) optimizer was used. The YOLOv3-416 model was used for wound bounding box detection. In SSD, we used the Pascal VOC annotations, which were converted to Pascal VOC format for the model training. The SSD model was trained for 475 epochs with a batch size of 8. The stochastic gradient descent (SGD) optimizer was used with a learning rate of 0.001. The SSD300 model is used for our detection. All the models were written in Python programming language by using the Pytorch deep learning framework and trained on a Nvidia GeForce RTX 2080Ti GPU platform.

III. MOBILE APPLICATION

At the start of the iOS-based application, the user is greeted with a login interface. Once the user is authenticated, application opens the live camera feed view, showing whatever is in view of the camera at a given moment as demonstrated in Fig. 3. (a). The code follows the implementation from Moonl1ght [22]. To detect objects (wound ROI), we use the YOLOv3 architecture as discussed above and the CoreML framework. For this application, we have used our model that is trained using approximately 4,000 images. Threshold IoU and object confidence is set to 0.5 and 0.2 respectively. Images shown in Fig. 3. (b)., are the screenshot of our iOS application detecting wound ROI.

Our application supports two main functionalities: 1) allowing the user to take a picture with the device camera and detecting the ROI in the taken picture, and 2) allowing the user to detect ROI in a live video mode. These requirements are met by three UIViewControllers, namely: OnlineViewController, PhotoViewController and SettingsViewController. The first is an online ROI detection for each frame. One may take a picture in the second, or pick a picture from the list, and check the network on those images. The third contains the settings: one can choose the model YOLOv3 or YOLOv3-tiny, as well as the thresholds.

A. *Core ML Framework*

Core ML is an Apple-developed machine-learning system, which is the basis for the features and domain-specific frameworks. Core ML builds on top of low-level primitives such as Accelerate and BNNS as well as Metal Quality Shaders themselves [23]. It is available for iOS 11 and above versions. The Core ML framework provides a unified representation for all models. By leveraging the CPU, GPU, and Neural Engine, Core ML optimizes on-device performance while reducing its memory footprint and power consumption. Running a model directly on the device eliminates the need for a network connection, which helps to keep the data of the user private and sensitive to the app.

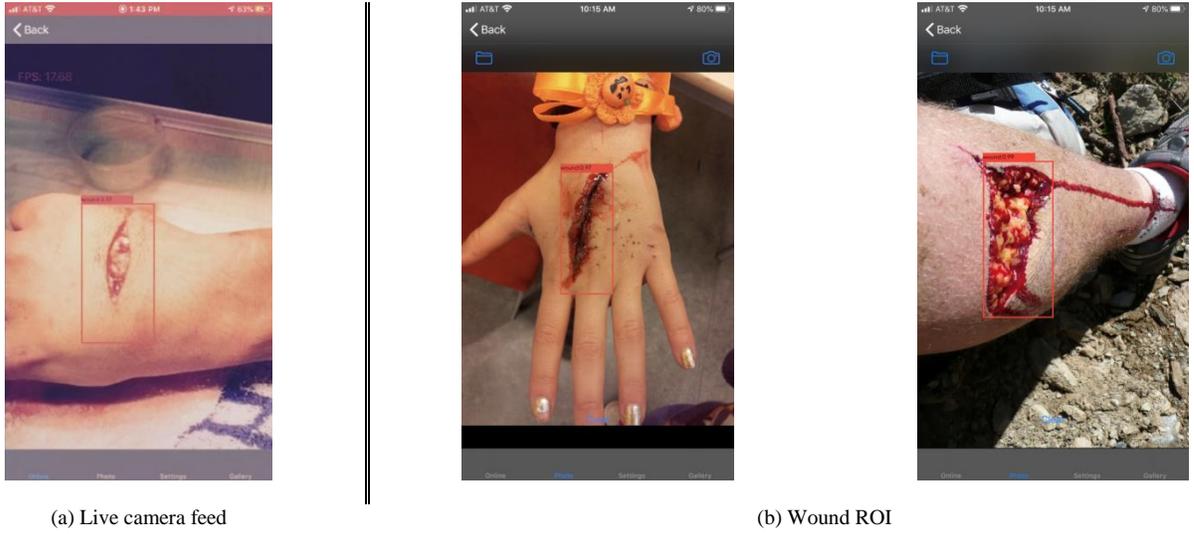

(a) Live camera feed      (b) Wound ROI

Fig. 3. Mobile Application.

### B. Implementation

CoreML only understands the .coreml model's unique format. For most common libraries, such as Tensorflow, Keras, or XGBoost, direct conversion to CoreML format is possible, but yet there is no such support in Darknet. We use YOLOv3 implementing Keras by following the following steps: loading the Darknet weights into the Keras model, saving it in the Keras format, and then converting it directly to CoreML. Since we are using pytorch to train the YOLOv3 model, we convert pytorch model to Darknet weights. After converting the trained model from Darknet format to CoreML, we have a file with the .mlmodel extension. YOLO class is primarily responsible for loading files from .mlmodel and managing outputs from the model.

We have three output layers for YOLOv3-416 and two for YOLOv3-tiny where bounding boxes for various objects are predicted. The YOLOv3 model uses three layers as output to break the image into a separate row, with grid cell sizes of 8, 16, and 32 pixels. Assuming an image in size of 416x416 pixels, the output matrices will be 52x52, 26x26 and 13x13 respectively. In the case of YOLO-tiny, all is the same, but we have two instead of three grids: 16 and 32, that is, 26x26 and 13x13 dimensional matrices. After the loaded CoreML model is started we get two (or three) MLMultiArray class objects on the display.

Once we receive the coordinates and sizes of bounding boxes and the corresponding probabilities for all found objects in the image, we can start drawing them on top of the image. We use a simple algorithm called non maximum suppression to eliminate the redundant boxes thereby reducing the complication of when one entity or object is expected to have many boxes with very large probabilities.

## IV. RESULT AND DISCUSSION

### A. Performance Metrics

We have adopted Precision, Recall, F1 score, Intersection over Union (IoU) and the Mean Average Precision (mAP) as the evaluation metrics to evaluate the localization performance. A brief description of these metrics is given below:

#### 1) Intersection over Union (IoU)

Intersection over union measures the overlap between the ground truth box (manually localized wound with labelImg) and the predicted box (model result), over their union. The IoU is calculated with the equation (1).

$$IoU = \frac{Ground\ Truth\ Box\ \cap\ Predicted\ Box}{Ground\ Truth\ Box\ \cap\ Predicted\ Box} \quad (1)$$

#### 2) Precision, Recall, and F1 score

To define the Precision, Recall and F1 scores, we set a threshold of IoU to 0.5. If IoU > 0.5, the result is said to be true positive. If IoU < 0.5, the result is false positive. If IoU > 0.5 and wound is wrongly classified, then the result is false negative. Precision and Recall show the accuracy of our

$$Precision = \frac{True\ Positive}{True\ Positive + False\ Positive} \quad (2)$$

$$Recall = \frac{True\ Positive}{True\ Positive + False\ Negative} \quad (3)$$

$$F1 = 2 \times \frac{precision \times recall}{precision + recall} \quad (4)$$

localization. Precision measures the percentage of correctly localized images in the wound localization, while Recall measures the percentage of correctly localized images in the ground truth. F1 score is the weighted average of precision and recall. Higher F1 score indicates better performance. Equations (2), (3), and (4) show the definitions of Precision, Recall, and F1 score metrics respectively.

### 3) Mean Average Precision (mAP)

The mean average precision compares different object detectors over multiple datasets. mAP calculation requires interpolated precision which is simply the highest precision value for a specific recall value. Interpolated Precision is calculated by using Equation (5). mAP is calculated from the summation of interpolated precision values as shown in Equation (6).

$$P_{interp}(r) = \max_{r' \geq r} p(r') \quad (5)$$

$$mAP = \sum_{r=0}^{1}(r_n - r_{n-1})P_{internp}(r_n) \quad (6)$$

### B. Result

We have tested the performance of both of our models with the test set of 405 wound images. For YOLOv3, with IoU of 0.5 and non-maximum suppression of 1.00, we get the mAP value of 0.939. The precision, recall and f1 score of this model is 0.925, 0.905, and 0.915 respectively. In SSD, by using an IoU of 0.5 and non-maximum suppression of 0.45, we get the mAP value of 0.864. Precision, recall and f1 score of SSD model is 0.902, 0.584, and 0.709 respectively. For mobile application, we have also implemented a light version of YOLOv3 (with reduced convolutional layers), named tiny-YOLOv3. With IoU threshold set to 0.5, we get the mAP value of 0.926 for tiny-YOLOv3. The precision, recall and f1 score of this model is 0.902, 0.899, and 0.9 respectively. Table I shows a brief summary of our evaluation results.

The robustness and reliability testing on Medetec dataset show very promising result. With our best model (YOLOv3 according to our AZH Wound Database evaluation), the precision, recall, f1-score, and mAP values are 0.926, 0.603, 0.73 and 0.808 respectively. Some of the testing outputs with YOLOv3 and SSD models are shown in Fig. 4.

### C. Discussion

From the results shown above, it is clear that YOLOv3 gives significantly better results. The mAP value of YOLOv3 is much higher than that of the SSD model with a difference of 7.5%. All the evaluation metrics (precision, recall and f1-score) reflect better values for YOLOv3 than SSD model. From Table I, we can see that SSD reflects a low recall and high precision, which leads to the decision that SSD is a very picky or fault-finding model. Most images detected as wounds are true wounds, but it also misses a lot of actual wounds. On the other hand, Table I also tells that YOLOv3 has a high precision (0.925) and high recall (0.905) value, representing a better and stable model.

Table I also indicates that the tiny-YOLOv3 gives very good results, in addition to its high speed. All the mAP, recall, precision, and f1-score values are better than SSD, and close to YOLOv3 results. For mobile platforms, it is critical to weight high on the lightness and speed of tiny-YOLOv3, while still achieving considerably high detection scores.

Regarding the robustness and reliability test, a promising result with a mAP value of 0.808 is achieved by our YOLOv3 model. With our model trained on AZH Wound Database and Medetec being a completely new and unseen dataset, this evaluation result is reasonable. From Fig. 4. we can see that YOLOv3 produces better results than the SSD model. In Fig. 4., from (a), (b), and (c) we can see that the SSD model misses some wounds in multiple wounds in a single image case, which shows the picky behavior of the SSD model as discussed above. From 4(d) and 4(e), we can see that, both models do good job; but SSD captures slightly more healthy skins than YOLOv3. So, we can confidently say that, YOLOv3 does a good job for wound localization than the SDD model.

Both YOLOv3 and Tiny-YOLOv3 models perform better than Goyal et al.'s wound localization work [25], where they achieved a mAP value of 0.849, 0.872, 0.918, and 0.906 for SSD-MobileNet, SSD-InceptionV2, Faster R-CNN with InceptionV2, and R-FCN with Resnet 101 models respectively. Their dataset contains only diabetic foot ulcer images (1775), but our dataset contains all types of ulcer images (1010), and this comparison may vary depending on the dataset. As the dataset of [25] is not publicly accessible, we implemented their best model (Faster R-CNN with InceptionV2) on our AZH Wound Database. This model uses Inception V2 for feature extraction and Faster R-CNN for object localization. Fig. 5. shows the comparison of Faster R-CNN with InceptionV2 model with our YOLOv3 and SSD models. From this figure, our model performs better than Goyal et al.'s best model on our dataset. In general, Darknet-53 (used by YOLOv3 for feature extraction) is much newer and better [26] than InceptionV2 [27] and the same claim goes for YOLO layers against Faster RCNN layers [28], which is clearly reflected on Fig. 5.

TABLE I
RESULT SUMMARY

| Network | Precision | Recall | F1-Score | mAP |
| --- | --- | --- | --- | --- |
| YOLOv3 | 0.925 | 0.905 | 0.915 | 0.939 |
| SSD | 0.902 | 0.584 | 0.709 | 0.864 |
| Tiny-YOLOv3 | 0.902 | 0.899 | 0.9 | 0.926 |

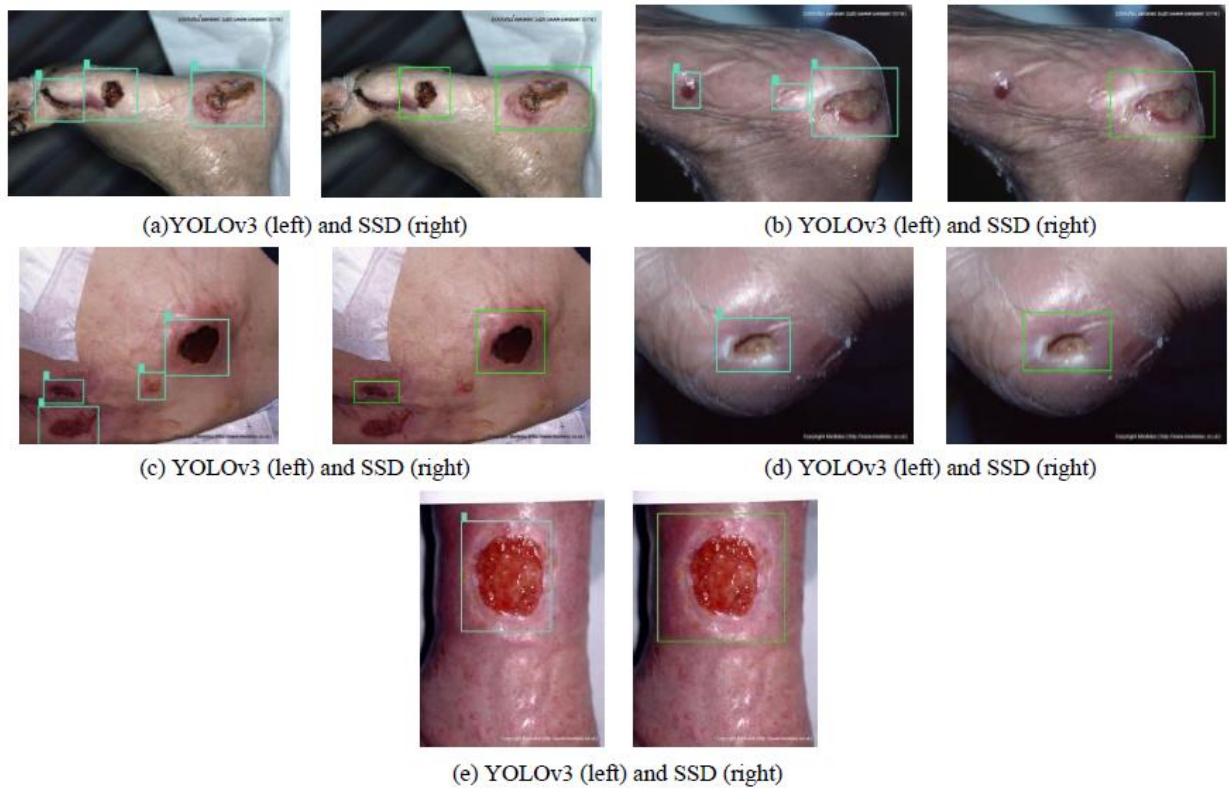

Fig. 4. Robustness and Reliability Testing Output.

## V. Conclusion

This research is focused on building an automated wound localizer, which is the first step of building an intelligent wound diagnostic system. The output of the localizer will be the input of subsequent wound processing tasks, such as wound segmentation and classification. Goyal et al. [25], recently demonstrated wound localization from diabetic foot ulcer images and achieved a highest mAP value of 0.918 by using Faster R-CNN with InceptionV2 model. Our system achieves a maximum mAP value of 0.939 and outperforms the only existing automated wound localization work. We have automated our wound localizer which is unique compared to most of the previous works based on localizing wounds manually from the original image. The present system has great importance in future research of intelligent wound healing. We have further broadened the scope of our work by building a mobile platform for it. By incorporating our technology into a smartphone platform, we hope to enhance patient access and wound care management strategies, improve clinical outcomes and provide cost effectiveness. Future work will include integrating wound segmentation and classification into the current wound localization platform on mobile devices.
.

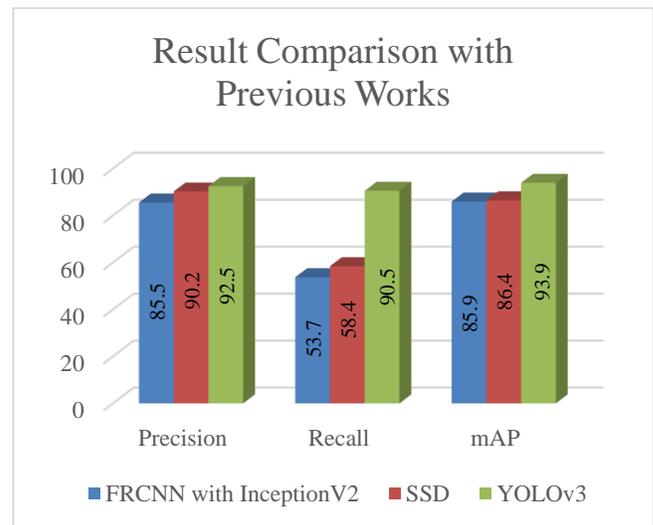

Fig. 5. The Faster R-CNN with InceptionV2 (Best Model Investigated in [25]), Compared with the Two Models (YOLOv3 and SSD) We Employ in Our Work. All Three Models are Trained and Tested on the Same Dataset, the AZH Wound Database.